\documentclass{article}

\usepackage{amsmath}
\usepackage{caption}
\usepackage{arxiv}
\usepackage{graphicx}
\usepackage[utf8]{inputenc} 
\usepackage[T1]{fontenc}    
\usepackage{hyperref}       
\usepackage{url}            
\usepackage{booktabs}       
\usepackage{amsfonts}       
\usepackage{nicefrac}       
\usepackage{microtype}      
\usepackage{lipsum}
\usepackage{multicol}
\usepackage{multirow}
\usepackage{subcaption}
\usepackage{arxiv}
\usepackage[square, numbers, comma, sort&compress]{natbib}
\usepackage{xcolor}
\usepackage{natbib}
\usepackage{graphicx}
\usepackage{color,soul}

\usepackage{hyperref}
\usepackage{algorithm}
\usepackage{algpseudocode}
\usepackage{algorithmicx}
\title{On Reward Shaping for Mobile Robot Navigation: A Reinforcement Learning and SLAM Based Approach}

\author{Nicolò Botteghi \\
  Robotics and Mechatronics (RaM)\\
  University of Twente\\
  Enschede, The Netherlands \\
  \texttt{n.botteghi@utwente.nl} \\
   \And
   Beril Sirmacek \\
   J\"{o}nk\"{o}ping AI Lab (JAIL)\\
   School of Engineering\\
   J\"{o}nk\"{o}ping University, Sweden\\
   \texttt{beril.sirmacek@ju.se} \\
   \And
   Khaled A. A. Mustafa \\
   Robotics and Mechatronics (RaM)\\
   University  of Twente \\
   Enschede, The Netherlands \\
   \And
   Mannes Poel \\
   Datamanagement and Biometrics (DMB) \\
   University of Twente \\ 
   Enschede, The Netherlands \\
   \And
   Stefano Stramigioli \\
   Robotics and Mechatronics (RaM)\\
   University of Twente \\
   Enschede, The Netherlands \\
}

\begin{document}
\maketitle

\begin{abstract}
We present a map-less path planning algorithm based on Deep Reinforcement Learning (DRL) for mobile robots navigating in unknown environment that only relies on 40-dimensional raw laser data and odometry information. The planner is trained using a reward function shaped based on the online knowledge of the map of the training environment, obtained using grid-based Rao-Blackwellized particle filter, in an attempt to enhance the obstacle awareness of the agent. The agent is trained in a complex simulated environment and evaluated in two unseen ones. We show that the policy trained using the introduced reward function not only outperforms standard reward functions in terms of convergence speed, by a reduction of 36.9\% of the iteration steps, and reduction of the collision samples, but it also drastically improves the behaviour of the agent in unseen environments, respectively by 23\% in a simpler workspace and by 45\% in a more clustered one. Furthermore, the policy trained in the simulation environment can be directly and successfully transferred to the real robot. A video of our experiments can be found at: \url{https://youtu.be/UEV7W6e6ZqI}
\end{abstract}

\keywords{Computer vision \and SLAM}

\section{Introduction}
The autonomous robot navigation problem is traditionally tackled by path planning algorithms (i.e potential field, cell decomposition, \textit{A* graph search} \cite{stentz}) when complete knowledge about the environment does exist. In most cases, this knowledge corresponds to the knowledge of the map of the environment in which the robot has to navigate in. These maps are usually built using Simultaneous Localization and Mapping (SLAM) algorithms \cite{Probabilistic}. However, in many interesting applications, full representation of the environment is expensive to obtain or difficult to keep up-to-date.

In recent years, Reinforcement Learning (RL) and its combination with function approximators, Deep Reinforcement Learning (DRL), has been used to tackle and solve several robotics tasks such as stabilization, manipulation, locomotion and navigation. 

In \cite{Zhang}, a successor feature DQN based reinforcement learning is proposed to solve the navigation problem when a map of the environment is known a priori. The main focus was to transfer the knowledge from one environment to another where the input is depth images obtained through a kinetic sensor and the output is four discrete actions for robot's navigation. In \cite{Brunner}, Asynchronous Advantage Actor-Critic (A3C) approach was proposed to help a robot moving out of a random maze for which a map is given. The input to the system includes 2D map of the environment, the robot's heading and the previous estimated pose whereas the output is the navigation actions such as move (forward, backward, right and left). Furthermore, in \cite{Zhangb}, an external memory acting as an internal representation of the environment for the agent is fed as an input to a deep reinforcement learning algorithm. In this way, the agent is guided to make informative planning decisions to effectively explore new environments. On the other hand, the work presented in this paper is built upon the approach introduced in \cite{Tai} where a map-less navigation is proposed based on an asynchronous deep deterministic policy gradient algorithm.

The purpose of this research is to design a motion-planner using deep reinforcement learning that uses raw sensory data from the robot's on-board sensors to determine a sequence of continuous velocity commands (linear and angular), that the robot has to execute in order to reach goal locations in an environment with unknown obstacle configuration. For this reason, a model-free actor-critic algorithm based on deep deterministic policy gradient (DDPG) is chosen to solve the navigation problem of mobile robots. 
 
Here, it should be pointed out that the deep-RL algorithm represents the \textit{high-level controller} of the robot. Once the robot's navigation actions are determined, the robot's low-level controller executes each action by sending the appropriate commands to the actuators. 
 
The main novelty of the paper lies into shaping the reward function based on the online-acquired knowledge about the environment that the robot gains during training.  This knowledge is obtained through a grid-based mapping with Rao-Blackwellized particle filter approach. In this way, the robot can learn a (sub)optimal policy in less number of iteration steps by increasing its awareness about the locations of the surrounding obstacles. Furthermore, the agent trained using this map-dependent reward function shows better generalization properties in case of complex and unseen environments than a standard RL map-less path planner. It is worth to mention that no map is used during the evaluation of the RL path planner in unknown environments, thus our method can be considered a map-less path planner, since a single map is used only during the training phase.  To the best of the authors' knowledge, this is the first time that a grid-based map is combined with reinforcement learning to shape the reward function, during training, for mobile robot navigation. Furthermore, the learned planner can generalize to unseen virtual environments as well as to a real non-holonomic differential robot platform without any fine-tuning or training using real-world samples. This work extends a recent publication \cite{botteghi1}, where a similar reward function is introduced for improving the convergence rate of the RL algorithm in a simple simulation environment with a single target goal. 

The rest of the paper is organized as follows. In section II, the theory behind DDPG and Rao-Blackwellized particle filter is presented. Section III presents the motion planner design, while section IV describes the experiments performed. Furthermore, sections V and VI discuss the results and the conclusions. 

\section{BACKGROUND}
\subsection{Reinforcement Learning}
\label{DDPG-sec}
The navigation problem of mobile robots in unknown environments can be formulated as a reinforcement learning problem in which a (sub)optimal path is obtained through a trial and error interaction between the agent and the environment. The interaction process can be modelled as a Markov Decision Process (MDP) \cite{Kai}, $\mathcal{M} = \left(\mathcal{S}, \mathcal{A}, \mathcal{P}, \mathcal{R}, \gamma  \right)$, where $\mathcal{S}$ is the state space, $\mathcal{A}$ is a set of actions, $\mathcal{P}(s_{t+1}|s_t,a_t)$ is the probability transition distribution, $\mathcal{R}(s_t,a_t)$ is the reward function based on the state-action pair and $\gamma \in [0,1]$ is the discount factor. The aim of reinforcement learning is to find a parametric policy $\pi_\theta$, that maps states into actions, to maximize the expectation of the total sum of discounted rewards given as 
\begin{equation}
  J(\pi_\theta)= \mathbb{E}_{s \sim \rho_\pi}[Q(s,a|\theta^Q)|_{s=s_t,a=\pi(s_t|\theta^\pi)}]
  \label{return}
\end{equation}
where $\rho^\pi(s)$ is the state distribution under policy $\pi_\theta$, $s_t$ is given as the sensor's measurements and $a_t$ is the motion commands. The proposed approach in this paper is based on a model-free deep deterministic policy gradient algorithm within an \textit{actor-critic framework}. Thus, the parameters of the actor network, $\theta^\pi$, are adjusted in the direction of the gradient of the expected return, $\theta^\pi_{i+1}=\theta^\pi_i+\alpha_{\theta^\pi}\nabla_{\theta^\pi}J(\pi_\theta)$, in an attempt to maximize it. The gradient of the expected return is evaluated as
\begin{multline*}
    \nabla_{\theta^\pi}J(\pi_\theta)=\mathbb{E}_{s\sim \rho_\pi}[\nabla_a Q(s_t,\pi(s_t|\theta^\pi)|\theta^Q)\nabla_{\theta^\pi}\pi(s_t|\theta^\pi)]
    \label{return_gradient}
\end{multline*}
To evaluate the action-value function defined in equation (\ref{return}), the current state $s_t$ and action $a_t$ are given as inputs to the critic network that in turn evaluates the quality of taking this particular action from this certain state in the form of a Q-value. The aim of the critic network is to adjust its parameters, $\theta^Q$, in such a way that the loss function defined in (\ref{loss}) is minimized.
\begin{equation}
   L_i(\theta_i^Q) = \mathbb{E}_{s \sim \rho_{\pi}, a \sim \pi}[(Q(s_t,a_t|\theta_i^Q)-y_i)^2]
   \label{loss}
\end{equation}
where $y_i=r(s_t,a_t)+\gamma Q(s_{t+1},a_{t+1}|\theta_i^Q)$ and $\pi$ is the behavioral policy that the robot uses for exploration. Based on that the parameters of the critic network are adjusted in the direction of the gradient of the loss function $\theta_{i+1}^Q=\theta^Q_i+\alpha_{\theta^Q}L(\theta_i^Q)$.\\
As a matter of fact, both expectations in (\ref{return}) and (\ref{loss}) are estimated by averaging over a mini-batch of samples that is randomly extracted from a replay buffer $\mathcal{D}$ to break the correlation between samples and as a result improve the stability of training \cite{nature}.

\subsection{Simultaneous Localization and Mapping}

In order to increase the robot's situation awareness, a SLAM algorithm is introduced in this section. SLAM is a well investigated problem in the robotics community that depicts the problem of robots building maps of the environments they are located in while simultaneously estimating their pose \cite{Probabilistic}. 

In this paper, a Rao-Blackwellized particle filter is used. The Rao-Blackwellized method separates the estimation of the robot's pose from the posterior of the map \cite{Murphy} as shown in equation (\ref{Rao-Black}):
\begin{equation}
     p(x_{1:t},m|z_{1:t}, u_{1:t-1})=p(m|x_{1:t}, z_{1:t}) p(x_{1:t}|z_{1:t}, u_{t-1})
     \label{Rao-Black}
\end{equation}
where $x_{1:t}$ represents the robot's trajectory, $z_{1:t}$ is the set of observations, $u_{t-1}$ is the control input and $m$ defines the built map of the environment.\\The advantage of using this factorization is that the estimation of the joint posterior can be divided into two separate steps:
\subsubsection{Particle Filter Estimation}
The particle filter uses a non-parametric representation of the probability distribution to estimate the pose of the robot $p(x_{1:t}|z_{1:t},u_{t-1})$ through a finite set of weighted particles (\ref{particles}).
\begin{equation}
\mathcal{X}_t := \{ x_t^{(1)},x_t^{(2)},...,x_t^{(N)}\},
\label{particles}
\end{equation}
where each particle $x_t^{(n)}$ represents a hypothesis of what the true state may be at time step $t$. Moreover, the current generation of particles $\mathcal{X}_{t}$ is constructed recursively from the previous generation $\mathcal{X}_{t-1}$ by sampling from a proposal distribution, i.e. the probabilistic odometry motion model $p(x_{t}^{(n)}|x_{t-1}^{(n)},u_{t-1})$. Then, by incorporating the probabilistic observation model $p(z_t|x_t^{(n)})$, an individual importance weighting factor $w_t^{(n)}$ is assigned to each particle. After that, a resampling step takes place, the particles with less importance weights die out and thus, the filter converges to the correct estimate. In this work, selective resampling is used in the resampling step \cite{selectiveresempling}.
\subsubsection{Mapping with Known Poses} After estimating the pose of the robot from the first step, it is possible to estimate the posterior of the map $p(m|x_{1:t},z_{1:t})$, i.e. ''mapping with known poses''\cite{Moravec}. This is done by using occupancy grid mapping as a way to model the environment. In that sense, the environment is divided into evenly spaced cells where a probability distribution is assigned to each cell in the grid indicating whether it is occupied, free or unknown. As a matter of fact, the resolution of the grid cells should be compatible with the smallest feature of the environment that should be considered as an obstacle. In occupancy grid maps, it is assumed that the probability of every grid cell, whether it is occupied or not, is independent of each other. By making advantage of this assumption, the certainty of the estimation of the entire map can be
broken down into the problem of estimating the posterior of every grid cell $m_i$ in the map and then the posterior over the
entire map can be approximately estimated by:
\begin{equation}
    p(m|z_{1:t},x_{1:t}) = \prod_{i=0}^M p(m_i|z_{1:t},x_{1:t})
    \label{map:uncertainty}
\end{equation}
where $M$ represents the number of grid cells in the map.

\section{METHODOLOGY}
\subsection{Network Architecture}
The actor network represents the policy and thus it is responsible
for mapping the states into actions $a_t=\pi_\theta(s_t)$. The state vector, $ \mathcal{S}\in \mathbb{R}^{44}$, is selected to be composed by the observation from the
laser range finder, that can be represented as 40-dimensional laser
beams with $180^\circ$ field of view (FOV) $x_t$, the relative distance between the target and the agent represented in polar coordinates $p_t$ and finally the latest executed action by the agent $v_{t-1}$. The actor's neural network is composed of three fully-connected
hidden layers with 512 nodes each which are activated by a rectified linear unit (ReLU) activation function. The output of the
actor's network is a 2-dimensional vector representing, respectively, the linear
and angular velocities of the robot, $\mathcal{A} \in \mathbb{R}^2$. For this purpose,
a sigmoid activation function is used to constrain the linear motion of the robot in the range between [0,1]. Furthermore, to constrain the angular velocity of the robot within [-1,1], a hyperbolic tangent function ''tanh'' is employed. Moreover, the output of the actor network is further multiplied by hyperparameters to limit the robot's linear velocity to $0.25 \text{m}/\text{s}$ and the maximum angular velocity to $1 \text{rad}/\text{s}$. The actor outputs, thus the actions, are then sent to the low-level controller to control the motion of the robot's actuators.\\
On the other hand, the critic network estimates the Q-value of a
state-action pair and thus its input is composed by the state and the action predicted by the actor. Similarly to the architecture described by \cite{DDPG}, the actions are not included until the second layer in the critic network to force the network to learn representations from states alone first. Based on the output of the critic network, the weights of both the actor and critic
are updated accordingly.  Like the actor network, the hidden layers of the critic network are activated
by a ReLU function. Finally, the activation of the output layer of the critic network presents a linear function.

\raggedbottom 
\subsection{Reward Function}
The agent's goal is to reach a desired target in the least number of iteration steps while avoiding colliding with surrounding obstacles. Thus, the agent receives a penalty proportional to the negative exponential of the Euclidean distance between its current position and the goal position; the distance is computed as $d= {p_t^{x,y}-g}_2$ where ${p}_t$ represents the current position of the robot at time $t$ with respect to the inertial frame, and $g$ is the target's location in the robot's coordinate frame. In addition, a sparse reward is added if the agent reaches the target within a predefined tolerance and, likewise, a penalty if the robot either collides with an obstacle or exceeds the maximum  number of allowed iteration steps $T$ in a single episode without either reaching the target or hitting an obstacle. Based on the above considerations, the overall reward function $r(s_t,a_t)$ in shown in (\ref{reward}).

\begin{equation}
    r(s_t)=\begin{cases}
    r_{\text{reached}}, & d \leq d_{min},\\
    r_{\text{crashed}}, & s_{ts},  \\
    1-e^{\gamma d}, & \text{otherwise}.
    \end{cases}
    \label{reward}
\end{equation}
where $\gamma$ represents the decay rate of the exponential. 
\begin{figure}[H]
	\centering
	\includegraphics[width=0.45\textwidth]{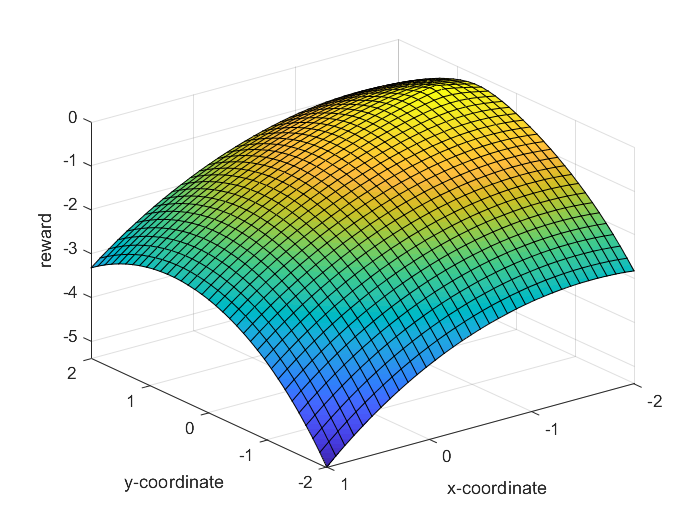}
	\captionsetup{justification=centering}
	\caption{A graphical representation of the dense reward function given in (\ref{reward}). The sparse penalty is not plotted for sake of clarity.}
    \label{fig:reward}
\end{figure}
A drawback of the dense reward function defined in (\ref{reward}) is that it only takes into account the target's location. In other words, even if the obstacles configuration changes, the \hbox{manifold} of the reward function will stay exactly the same. This means that the robot learns no awareness about obstacles inside the environment, but only their location by crashing against them, due to the sparse penalty. Therefore, to improve the learning performance and obstacle awareness, the reward function is shaped based on the acquired 2D occupancy grid map that the robot builds during training. This map is built by dividing the workspace in which the robot operates into evenly-spaced cells.  Every cell inside the occupancy grid is classified as occupied or free based on a predefined threshold value that
determines the occupation probability of each cell. This probability is updated while
the robot is exploring the environment. The incorporation of the environment's knowledge is weighted by the level of certainty of the map's posterior $p(m|z_{1:t},x_{1:t}) = \prod_{i=0}^M p(m_i|z_{1:t},x_{1:t})$. Moreover, since every obstacle inside the environment is represented by a certain number of occupied grid cells, this reward term is normalized by the total number of occupied grid cells in the field of view (FOV) of the robot. Intuitively, areas with more obstacle concentration will return more negative rewards, than free space assuming the same confidence in the map given by the map's posterior. This
can be formulated as follows:
\begin{equation}
    r_m(s_t, map)=\frac{1}{k}\prod_{i=0}^M p(m_i|z_{1:t},x_{1:t})\sum_{j=0}^k e^{-c_{j_{min}}},
    \label{slamreward}
\end{equation}
where $c_{j_{min}}$ represents the distance between the robot and an occupied cell in its field of view.
Now the reward function is not purely state dependent, but it depends on the map (distance to occupied cells), on time and on the sequence of laser measurements $z_{1:t}$ and poses $x_{1:t}$. Thus, (\ref{final_reward}) contains also history information. However, it is worth to mention that no map or history of measurements or poses are included in the state vector by partially violating the MDP model, since the reward is a function of the state. The choice is made because the final goal is to develop a map-less path planner. In our approach a single map  is built and used only during the training of the agent.\\
Eventually, the map dependent term defined in (\ref{slamreward}) is subtracted from the dense reward given in (\ref{reward}) where the sparse rewards are kept as they are.  The proposed reward function is presented in (\ref{final_reward}).
\begin{equation}
    r(s_t, map)=\begin{cases}
    r_{\text{reached}}, & d \leq d_{min},\\
    r_{\text{crashed}}, & s_{ts},  \\
    1-e^{\gamma d} - r_m(s_t, map), & \text{otherwise}.
    \end{cases}
    \label{final_reward}
\end{equation}

\begin{figure*}
        \centering
        \subcaptionbox{\label{fig3:a}}{\includegraphics[width=1.8in]{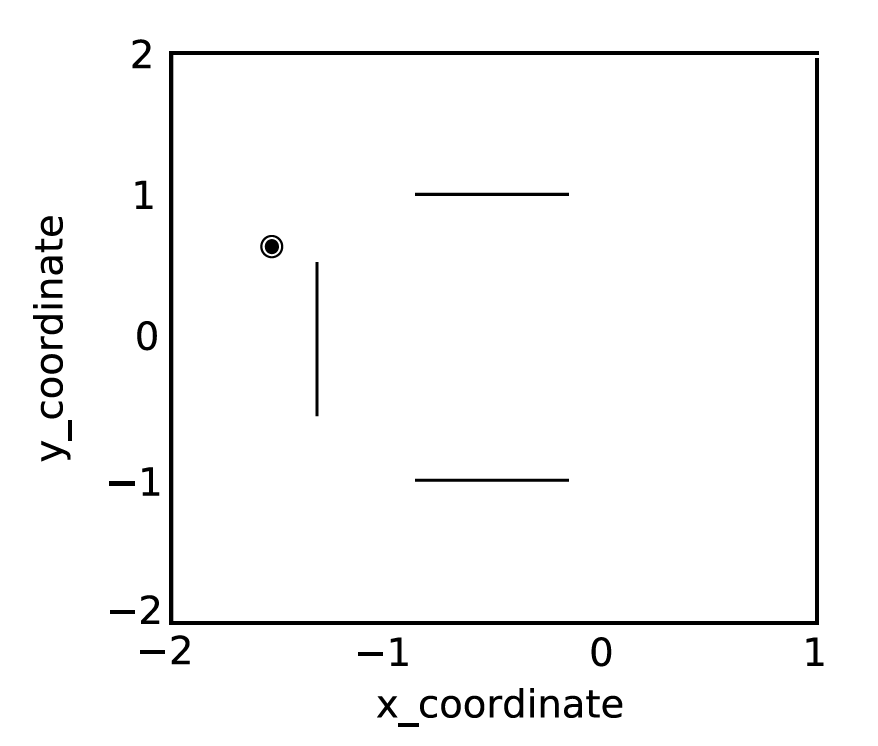}}\hspace{1em}%
        \subcaptionbox{\label{fig3:b}}{\includegraphics[width=2.2in]{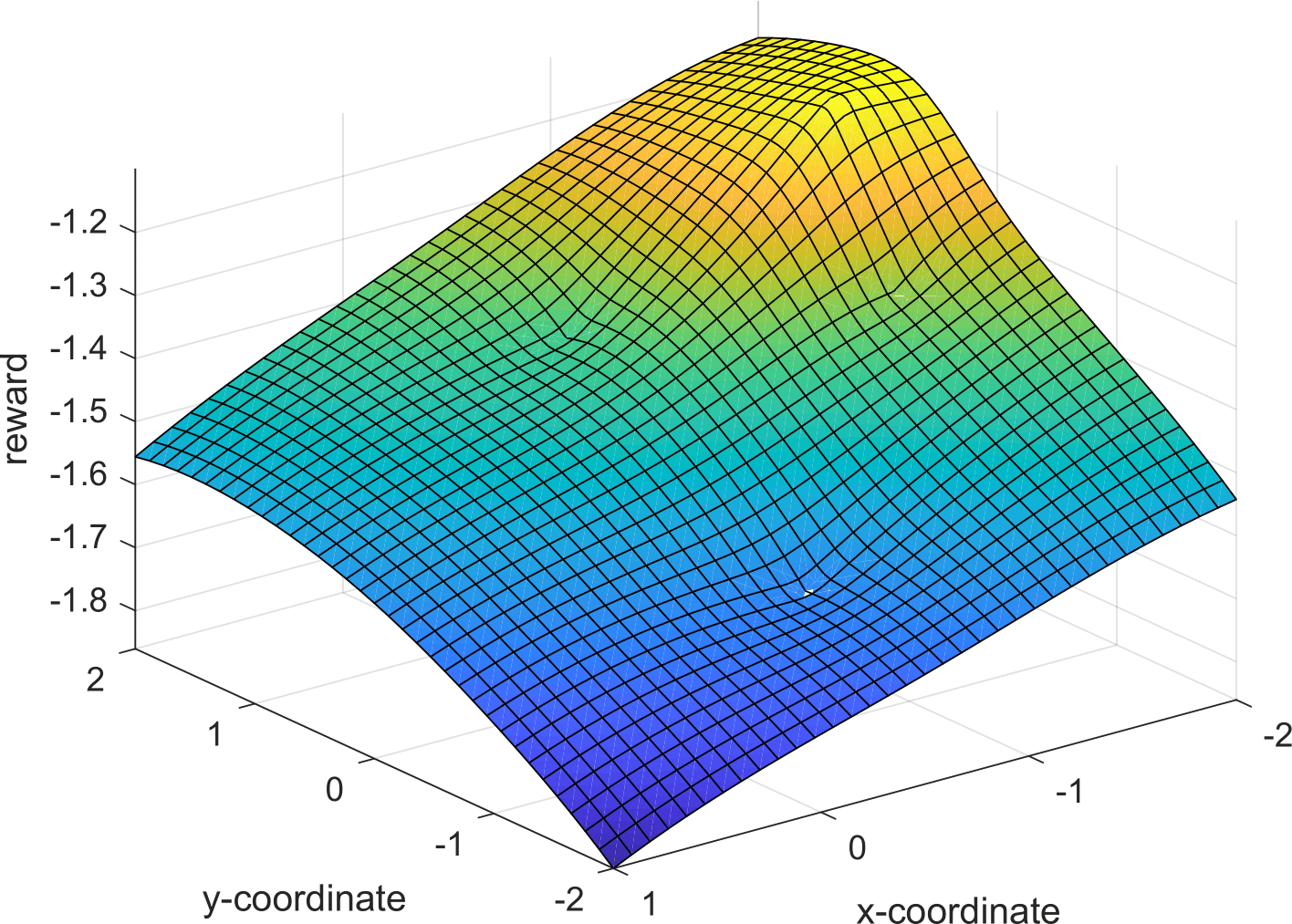}}\hspace{1em}%
        \subcaptionbox{\label{fig3:c}}{\includegraphics[width=2.2in]{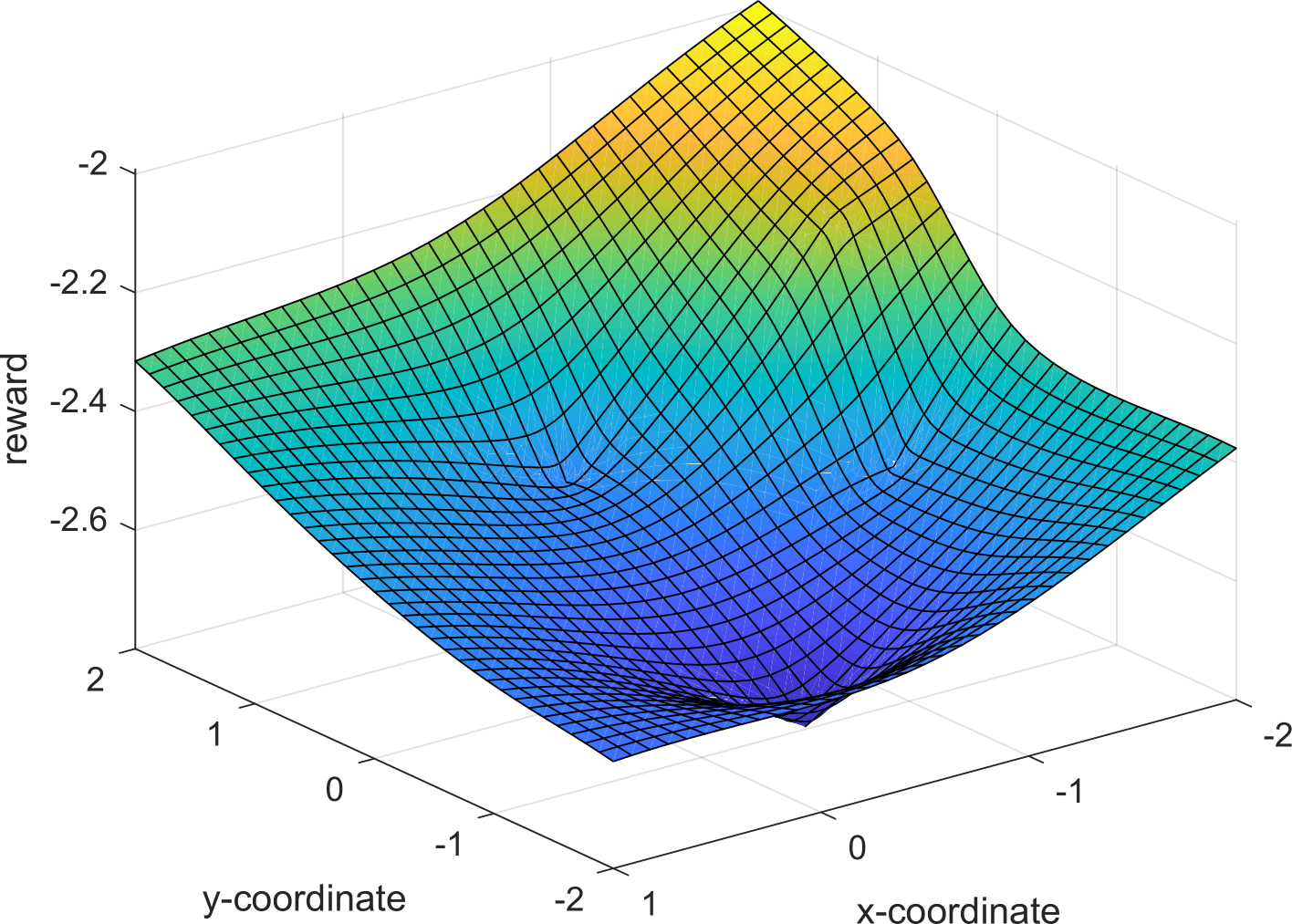}} 
        \caption{Training environment used to build the map. The goal is located at (-1.6,0.65) represented by the black circle in (a). The evolution of the effect of the map-dependent term on the reward function, when $p(m|z_{1:t},x_{1:t})$=0.5 in (b) and $p(m|z_{1:t},x_{1:t})$=1.0 in (c). It is possible to notice that the depressions of the reward function, thus the penalty received by the agents becomes more severe the more the map posterior grows.} \label{fig:rewevolution}
\end{figure*}

\noindent In this way, the reward function does not only depend on how far the agent is from the target, but on the distance to the obstacles inside the workspace. To visualize how the reward function varies with the change of the posterior of the map $p(m|z_{1:t},x_{1:t})$ that gradually increases as the robot becomes more confident about the map, the reward function (\ref{final_reward}) including the map-dependent term is plotted at two different instants as shown in Figure \ref{fig3:b} and \ref{fig3:c}. It is possible to notice three main depression areas corresponding to the three obstacles presents in the simple environment shown in Figure \ref{fig3:a}. 

\section{Experiments} 

\subsection{Experimental Setup}
The virtual 3D environment is built using the Gazebo simulator. The experiments were conducted on an Ubuntu 16.04 machine with an Intel Core i7-8550 CPU. The algorithm is implemented using OpenAI package provided by the Robot Operating System (ROS) middleware. In addition, the simulated platform is a skid-steering Husarion mobile robot. For training the model, stochastic policy gradient with Adam  optimizer \cite{Kingma} is employed to train both the actor and the critic networks. However, for the actor network, a learning rate of $10^{-4}$ is used whereas the critic is updated using a learning rate of $10^{-3}$.  Furthermore, L2 regularization is included with a coefficient of $10^{-2}$ when training the critic network to prevent overfitting. A discount factor of $\gamma = 0.99$ and target update, $\tau = 0.001$ is used. The initial values of weights and biases of the hidden neurons are sampled from a uniform distribution $[-\frac{1}{\sqrt{f}},\frac{1}{\sqrt{f}}]$ where $f$ is the number of inputs to the layer, while the values of weights and biases for the output layer are sampled from a uniform
distribution $[-3\times 10^{-3}, 3\times10^{-3}]$ to ensure that the outputs at the start of training are close
to zero. The hyperparameters are selected based on the ones used in the original paper for the DDPG \cite{DDPG}. The exploration noise is chosen as an Ornstein-Uhlenbeck process with parameters $\sigma = 0.2$ and $\theta = 0.15$, since these values have empirically shown the best performances.
The outputs of the policy are clipped to lie between the actuator limits after the addition of noise.  In all experiments, a standard replay buffer that holds up to  $100,000$ transitions is used, which means that, in the worst case, the buffer can store 100 episodes since the maximum number of iterations in every episode is limited to  $10^3$ time-steps. The updates of the weights of the networks are executed with a mini-batch of dimension 64. In order to  simultaneously map the environment and estimate the robot pose, the ROS gmapping SLAM package is used. The  inputs  for  mapping  includes  wheel  odometry, laser  range finder  data and the 2D occupancy grid map representing the environment. A probability value is assigned to each cell based on whether it is occupied or free according to the laser sensor and odometry readings. Similar to \cite{selectiveresempling}, the occupancy threshold value is chosen equal to 0.65  which means that if the probability value of the cell is greater than this value, this cell is occupied and, consequently, free otherwise. Besides that, to avoid the high computational load of the calculations, the map is only updated after certain change occurs to the probability of the posterior of the map $p(m|z_{1:t},x_{1:t})$ within a threshold of 0.25. The robot subscribes to laser readings with a scanning range from 0.2m to 2m.  The
position of the robot is evaluated through Rao-Blackwellized particle filter, instead of using raw
odometry data, in order to better estimate its location and more accurately compute the distance to the target. The agent is free to select any angular and linear velocities from
a continuous range as long as the values are feasible by the physical constraints of the robot. These
velocity commands are directly sent to the low-level controller, with frequency equal to 10 Hz, where the control loop waits until the command gets executed. This feedback is provided by estimating the robot’s velocity from
the encoder’s readings.

\subsection{Training Environment}
To validate the effectiveness of the proposed approach, the robot is trained on the virtual environment \textit{Env}-1 shown in Figure \ref{fig:sub1at}. To guarantee that the robot can reach any target inside the workspace without getting biased to certain trajectories, the desired target location $g$, at the beginning of each episode, is randomly sampled from a uniform distribution. In addition, the initial pose of the robot $\mathbf{p}_0$ is also randomly sampled from a uniform distribution in such a way that a collision-free path is guaranteed to exist between the initial pose of the robot and the desired target. In this way, it is ensured that the robot does not get biased by the structure of the environment on which it is trained. To further assess the ability of the learned policy to generalize to unseen environment, the agent is evaluate in both \textit{Env}-2 and \textit{Env}-3, Figures \ref{fig:sub2at} and \ref{fig:sub3at},  and the results are reported in section \ref{sec:eval}.


\section{RESULTS}
\subsection{Training Phase}
\begin{figure}[h] 
	\begin{center}
		\includegraphics[width=0.45\textwidth]{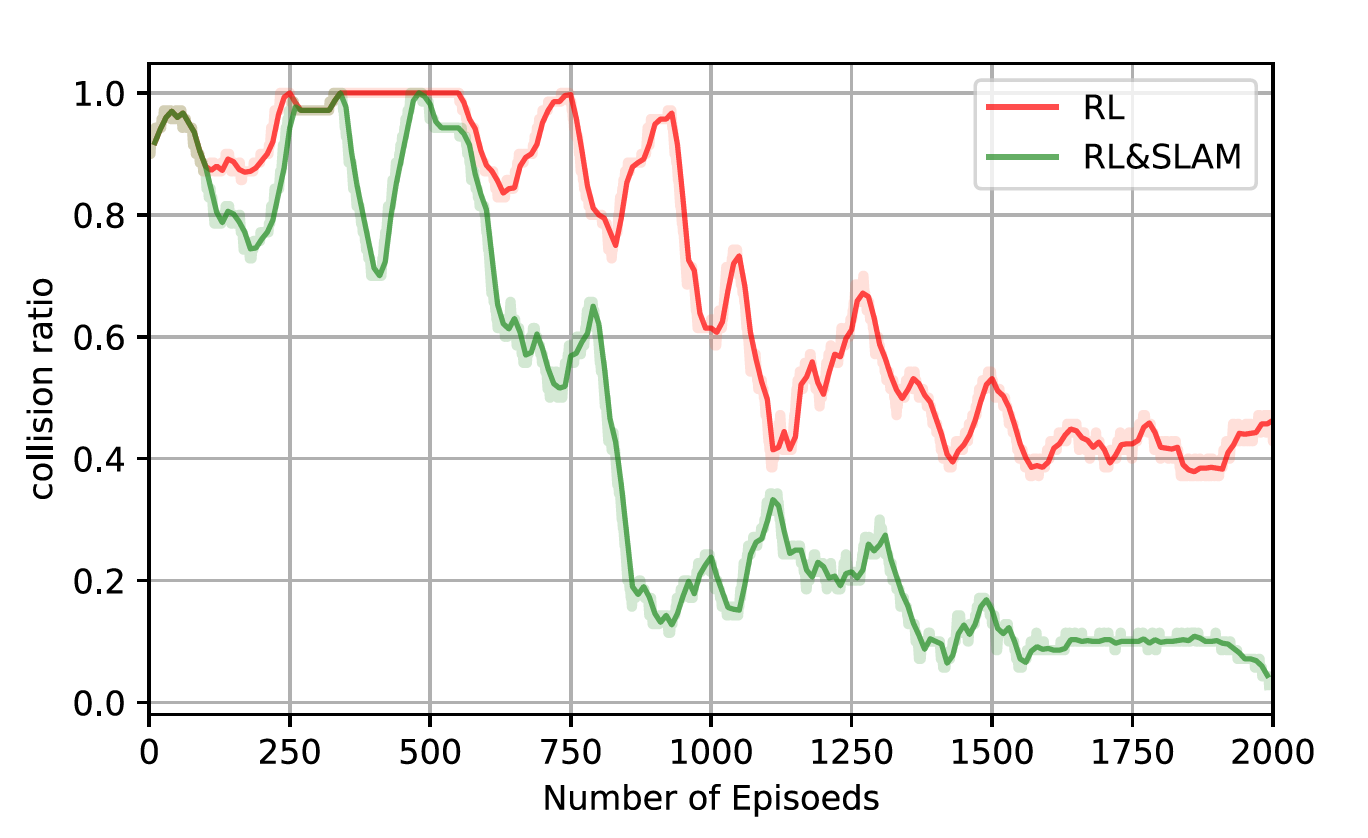}
	\end{center}
	\captionsetup{justification=centering}
	\caption{Evolution of the collision ratio with the number of training episodes. The collision samples of the proposed approach (green) decrease much faster than the standard approach (red).}
	\label{fig:collison_rate_exp1}
\end{figure}

\begin{figure*}[ht!]
\centering
\begin{subfigure}{.33\textwidth}
  \centering
  \includegraphics[width=0.55\linewidth]{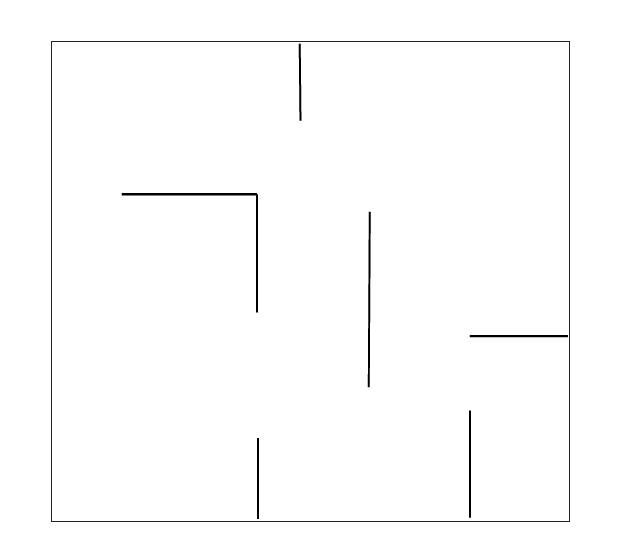}
  \captionsetup{justification=centering}
  \caption{\textit{Env}-1.}
  \label{fig:sub1at}
\end{subfigure}%
\begin{subfigure}{.33\textwidth}
  \centering
  \includegraphics[width=0.55\linewidth]{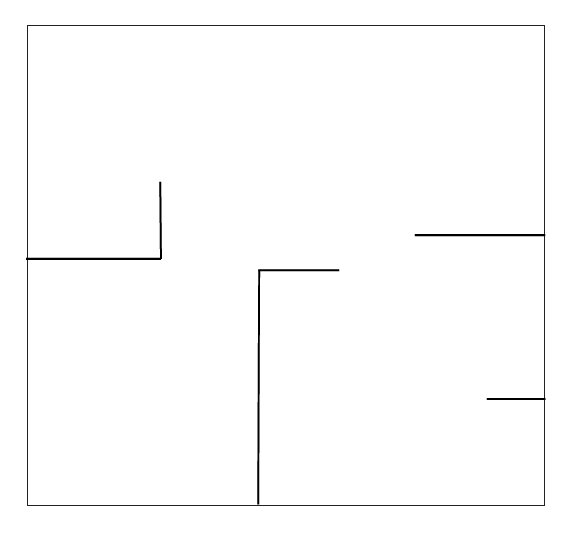}
  \captionsetup{justification=centering}
  \caption{\textit{Env}-2.}
  \label{fig:sub2at}
  \end{subfigure}%
  \begin{subfigure}{.33\textwidth}
  \centering
  \includegraphics[width=0.55\linewidth]{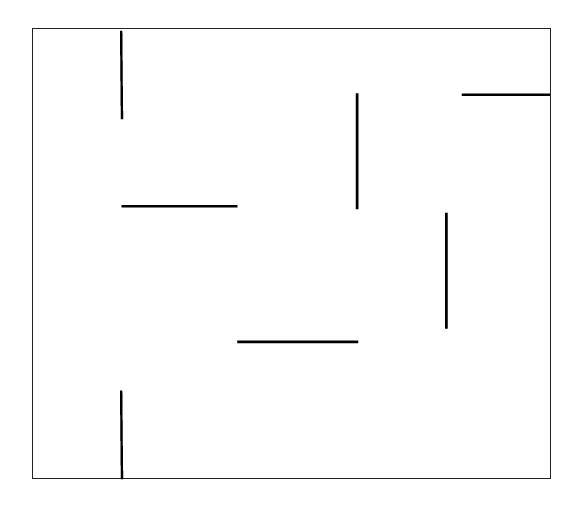}
  \captionsetup{justification=centering}
  \caption{\textit{Env}-3.}
  \label{fig:sub3at}
  \end{subfigure}%
\captionsetup{justification=centering}
\caption{The robot is trained on \textit{Env}-1 depicted in \ref{fig:sub1at}. Then, the performance of the learned policies are later evaluated on virtual environments 2 and 3 which were unseen to the robot during training. All environments are of size 5.5\text{m}x4\text{m}. }
\label{fig:envs}
\end{figure*}

In this section, the performance of both the standard reinforcement learning approach with reward function (\ref{reward}) and the one where the reward function is shaped based on the online knowledge of the map acquired during training (\ref{final_reward}) are evaluated. The evolution of the collision ratio with respect to the number of episodes is depicted in Figure \ref{fig:collison_rate_exp1}. At the beginning of training, both approaches have similar performances. However, after approximately 580 episodes, the robot has become more confident about the map it is building, and as a result, the probability term $p(m|z_{1:t},x_{1:t})$ has a higher impact on the shaping of the reward function.
Because during the training the target locations and the initial poses of the robot are sampled from a uniform distribution at the beginning of each episode, the curve in Figure \ref{fig:collison_rate_exp1} presents fluctuations. However, it is possible to notice that the fluctuations decrease the more the policy improves, since the agent can more easily reach targets independently of their position and its initial pose.
The agent trained with (\ref{final_reward}) achieves better understanding and awareness about the presence of the obstacles and thus, figures out collision-free paths to reach the desired targets in less number of episodes. Moreover, by the end of training, the proposed approach achieves a success ratio of 93\% compared to 68\% that is achieved by the standard RL approach (\ref{reward}). 
\noindent Furthermore, although both agents were trained for 2000 episodes, a 36.9\% reduction in the number of iteration steps required for convergence is shown by the agent trained with (\ref{final_reward}). These results show that incorporating the knowledge of the environment in the reward function dramatically outperforms the standard algorithm by drastically reducing the collisions during training and consequently improving the learning rate. 

\subsection{Evaluation Phase}

\label{sec:eval}
After the training phase, the learned policy is evaluated for both approaches on the same set of 100 random targets, to guarantee a fair comparison the percentage of the successfully solved episodes is monitored. Additionally, we compare with move\_base, the DWA path planner \cite{dwa_planner} in ROS. It is worth to mention that move\_base requires the map of the environment and several parameters need to be tuned to achieve good performances. An episode is considered to be successful, when the robot reaches the target without either hitting any of the obstacles or exceeding the maximum number of allowed actions per episode which is set to 300 steps during the evaluation phase. Furthermore, to investigate the effect of incorporating knowledge of the environment into the training process on the quality of the generated trajectories, the average number of actions required to reach these targets is recorded. First, the evaluation is performed on the same training environment \textit{Env}-1 and the results are reported in Table \ref{tab:env1}. Then, the policy is transferred to  the new and unseen environments \textit{Env}-2, \textit{Env}-3 and the results are illustrated in Table \ref{tab:eval}. As shown in Table \ref{tab:env1}, the proposed combined approach not only achieves a higher success ratio than the two other methods, but also tends to execute a much smaller average number of actions to reach the desired targets. This is another benefit of shaping the reward function using the introduced map-dependent term. Because of the enhanced awareness about the presence of the obstacles, the agent can figure out shorter, smoother and more efficient trajectories to the reach the desired targets.

\begin{table}[H]
  \begin{center}
  \captionsetup{justification=centering}
   \caption{Assessment of RL, RL\&SLAM and move\_base on the same training environment \textit{Env}-1.}
    \label{tab:env1}
    \begin{tabular}{l|c|c|c} 
    \toprule[\heavyrulewidth]\toprule[\heavyrulewidth]
       & approach & success ratio & number of actions\\
       & & $\%$ & (mean $\pm$ std)\\
      \hline
      \textit{Env}-1 & RL  & 68 \%  & 128.9$\pm$183.8 \\
      & RL \& SLAM & \textbf{93}
\% & \textbf{86.2} $\pm$ \textbf{68.1} \\ 
      & move\_base & 73 \% & 104.2$\pm$72.61 \\
\bottomrule 
    \end{tabular}
    \captionsetup{justification=centering}
  \end{center}
\end{table}



\noindent When transferred to \textit{Env}-2, \textit{Env}-3, the obstacle-aware policy trained on \hbox{\textit{Env}-1}, significantly improves the performance of the agent as it can be observed in terms of both success ratio and number of executed actions in Table \ref{tab:eval}. 
\begin{table}[H]
  \begin{center}
  \captionsetup{justification=centering}
   \caption{Assessment of RL, RL\&SLAM and move\_base on unseen virtual environments \textit{Env}-2, \textit{Env}-3.}
    \label{tab:eval}
    \begin{tabular}{l|c|c|c} 
    \toprule[\heavyrulewidth]\toprule[\heavyrulewidth]
       & approach & success ratio & number of actions\\
       & & $\%$ & (mean $\pm$ std)\\
      \hline
\textit{Env}-2 & RL  & 61 \%  & 149.7$\pm$208.9 \\
      & RL \& SLAM & \textbf{84}
\% & \textbf{85.5} $\pm$ \textbf{68.9}\\
& move\_base & 82 \% & 109.6$\pm$107.34 \\
\hline 
\textit{Env}-3 & RL  & 31 \%  &99.6$\pm$115.5 \\
      & RL \& SLAM & \textbf{76}
\% & 86.4 $\pm$101.2\\
& move\_base & 56 \%  & \textbf{75.2}$\pm$\textbf{62.15} \\
\bottomrule 
    \end{tabular}
    \captionsetup{justification=centering}
  \end{center}
\end{table}

\begin{figure}[h]
	\begin{center}
		\includegraphics[width=0.35\textwidth]{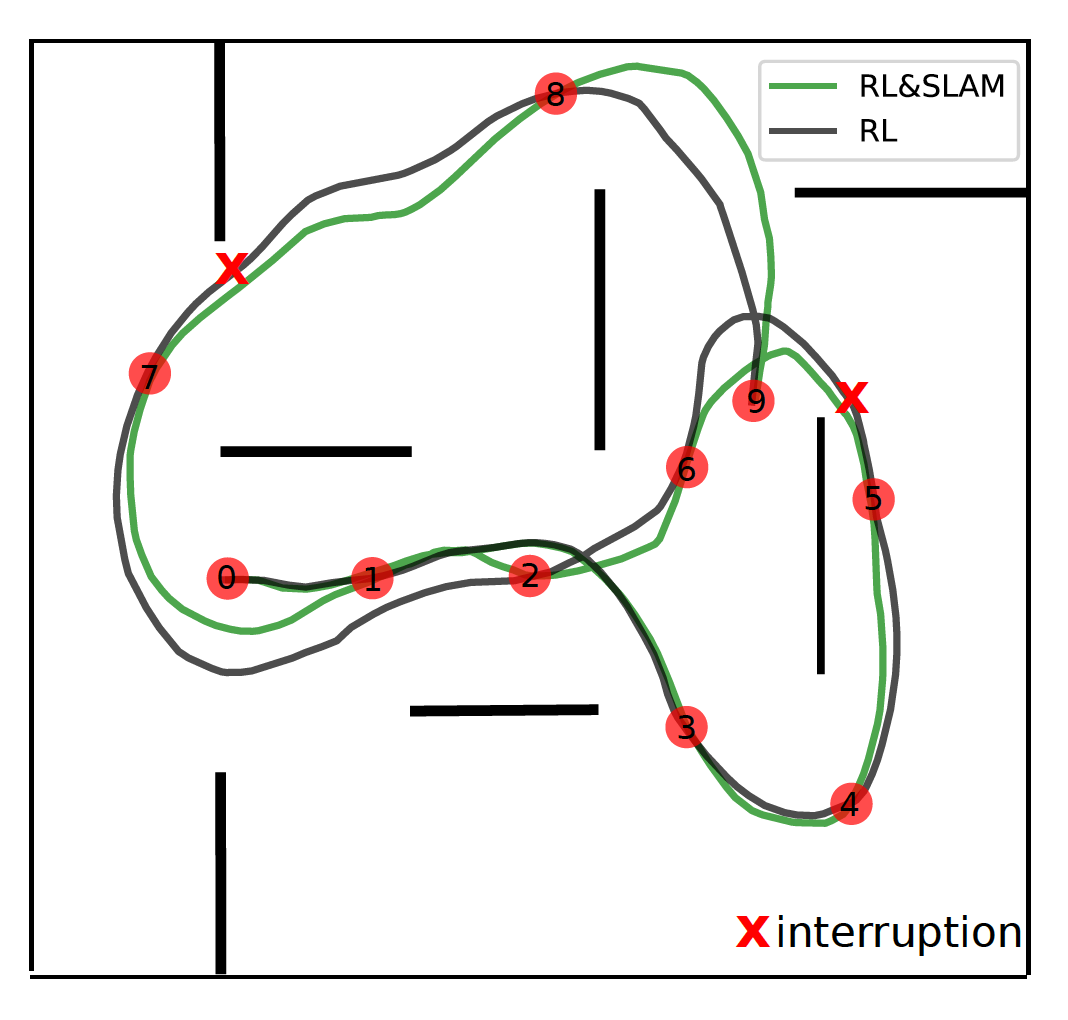}
	\end{center}
	\captionsetup{justification=centering}
	\caption{Trajectory tracking in the virtual environment \textit{Env}-3. The motion planner based on the
standard RL approach was not able to finish the navigation task, while the motion planner trained with the proposed reward function can reach all the targets without collisions.}
	\label{fig:traj}
\end{figure}

Even though the structure of \textit{Env}-2 is simpler and with less obstacles than the training environment \textit{Env}-1, the agent trained using the new reward function achieves a higher success ratio, 84\% against 61\%  and achieves performances slight superior to move\_base, but more interesting the average number of actions taken is drastically reduced by 64.2 and 24.1 respectively. In a more clustered environment as \textit{Env}-3 the new agent dramatically improves the success ratio, 76\% against 31\% and 56\% , while the reduction on the number of actions is mitigated, only 13.2 due to the fact that the standard RL planner often collides after executing few actions and exceeds by 11.2 the average number of actions of move\_base again due to the poor performances of the planner and the frequent collisions. Moreover, in Figure \ref{fig:traj}, the performances on the two agents in \textit{Env}-3 when a sequence of targets is given are shown. In both cases the proposed reward function (\ref{final_reward}) outperforms the reward function (\ref{reward}). 

This is due to the fact that, by incorporating knowledge of the environment in the reward function, the agent can better relate the features extracted from the laser data to the good or bad behaviours, respectively getting closer to the targets and getting closer to the obstacles. More importantly, the behaviours learned by the agent are independent of the training environment and this allows the agent to perform well on unseen environments with similar structures. On the other hand, for the standard approach (\ref{reward}), since the agent learns the positions of the obstacles only when it comes into their vicinity due to the sparse penalty, there is nothing that motivates the robot to choose safer paths and move away from the obstacles in advance resulting in much higher number of collisions. This indicates that the standard approach does not generalize well to more complicated and unseen environments. Furthermore, the proposed path planner achieves better performances than move\_base.

\raggedbottom
\begin{figure*}[ht!]
\centering
\begin{subfigure}{.33\textwidth}
  \centering
  \includegraphics[width=0.85\linewidth]{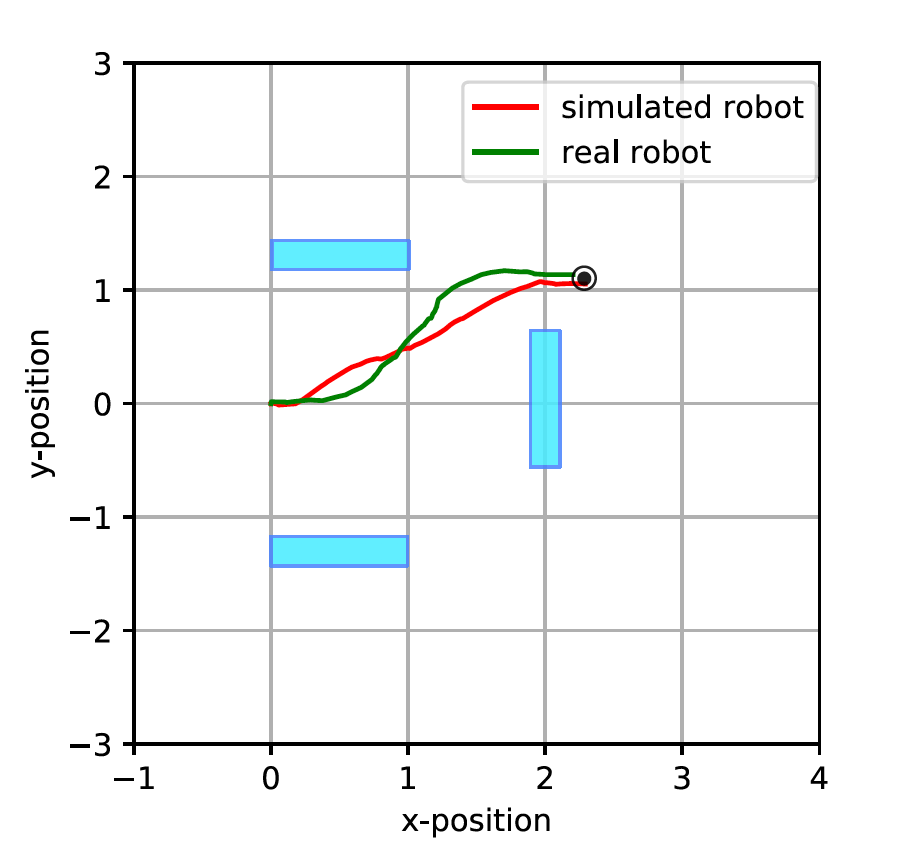}
  \captionsetup{justification=centering}
  \caption{\textit{Target}-1.}
  \label{fig:sub1a}
\end{subfigure}%
\begin{subfigure}{.33\textwidth}
  \centering
  \includegraphics[width=0.85\linewidth]{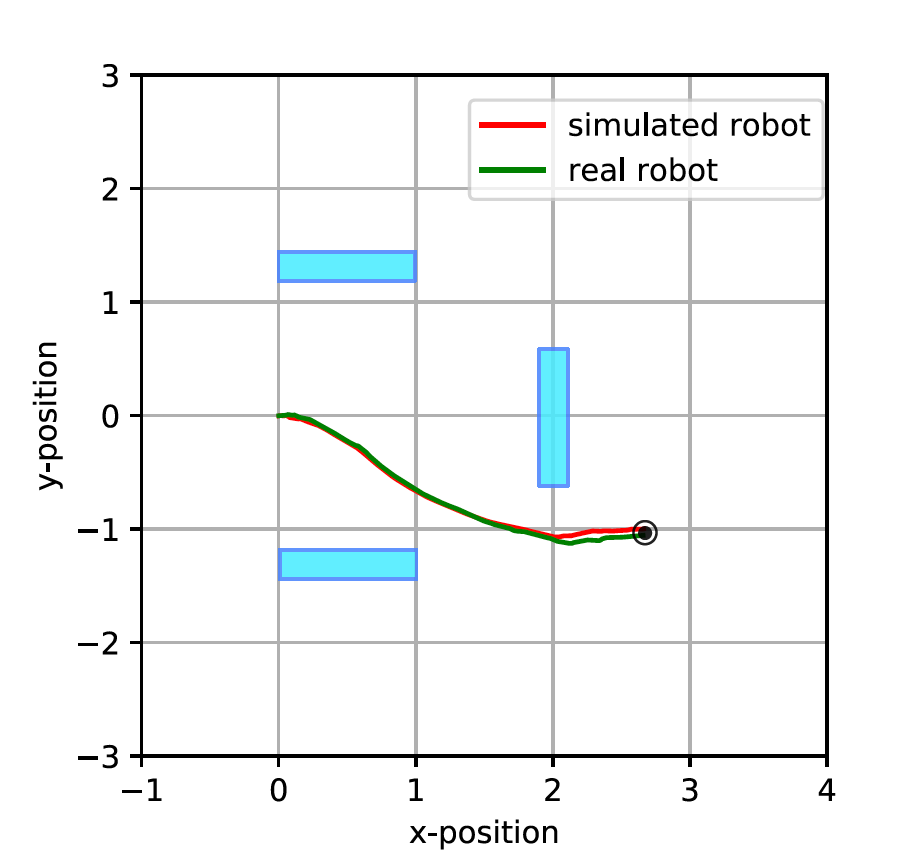}
  \captionsetup{justification=centering}
  \caption{\textit{Target}-2.}
  \label{fig:sub2a}
  \end{subfigure}%
  \begin{subfigure}{.33\textwidth}
  \centering
  \includegraphics[width=0.85\linewidth]{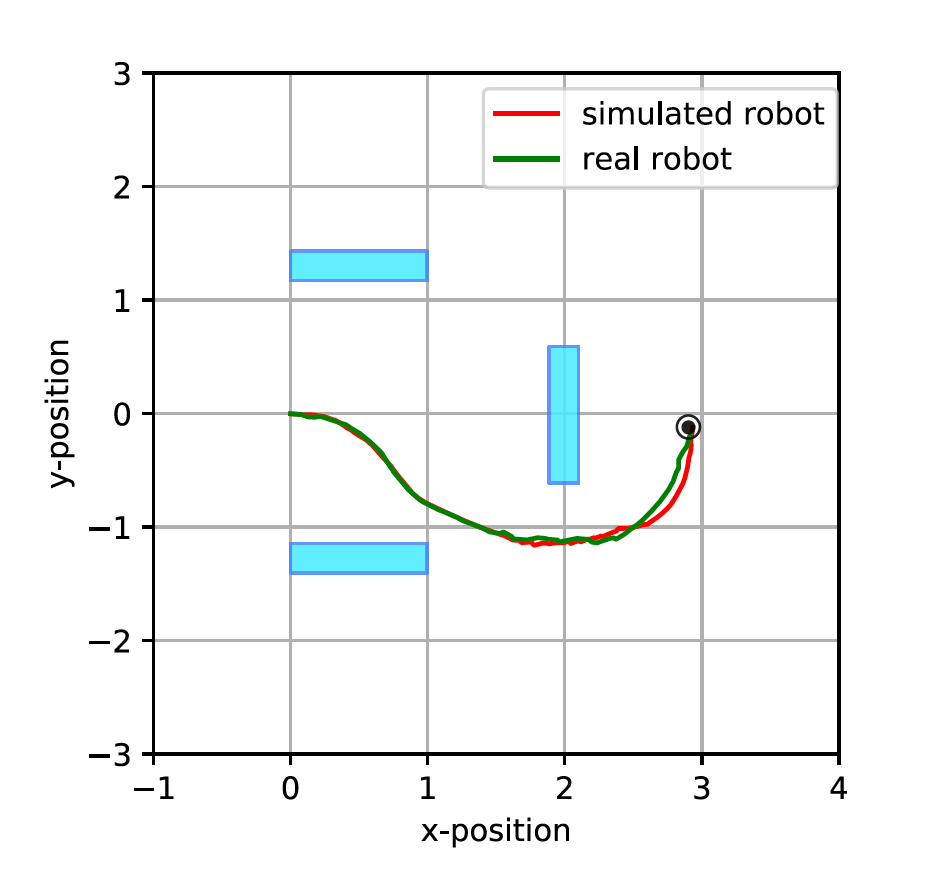}
  \captionsetup{justification=centering}
  \caption{\textit{Target}-3.}
  \label{fig:sub3a}
  \end{subfigure}%
\captionsetup{justification=centering}
\caption{A comparison of the generated trajectories by the simulated robot in gazebo (visualized in
red) and the real robot after transferring the learned policy (visualized in green). The target is depicted
by the black circle. }
\label{fig:real_robot}
\end{figure*}

\subsection{Real-world Experiments}

\begin{figure}[H]
	\begin{center}
		\includegraphics[width=0.35\textwidth]{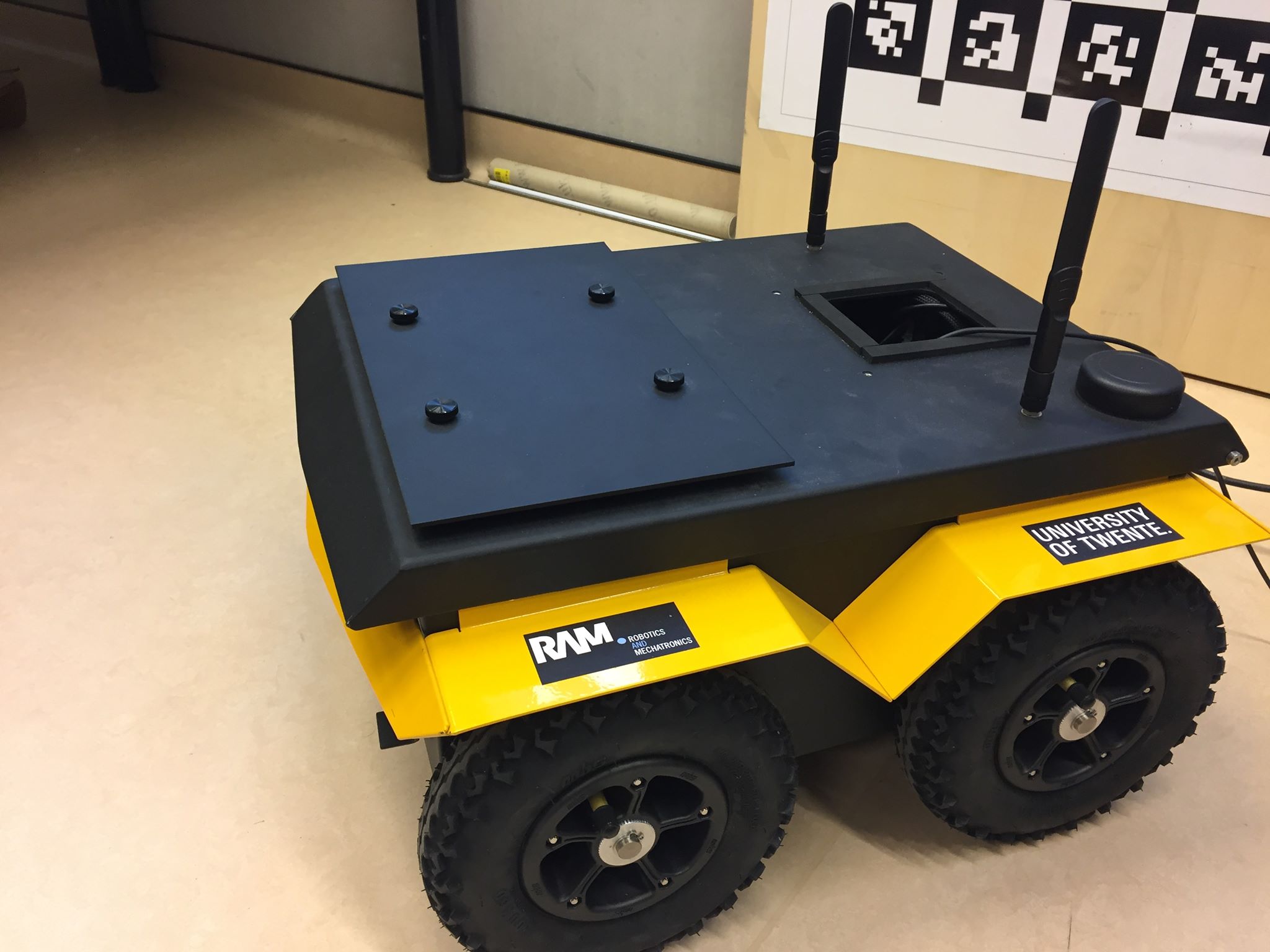}
	\end{center}
	\captionsetup{justification=centering}
	\caption{Clearpath Jackal differential drive mobile robot used in the real-world experiments at University of Twente. }
	\label{fig:struct}
\end{figure}

Obtaining real-data from robotic systems can be extremely difficult and time-consuming. For instance, for deep reinforcement learning algorithms that require huge number of samples for their convergence, e.g. $10^5$ samples as in the navigation problem under study. Obtaining those numbers of samples on a real robot is almost impossible. However, simulation with accurate models could potentially be used to offset the cost of real-world interactions. For the real-world experiments, a Clearpath Jackal differential drive mobile robot, Figure \ref{fig:struct}, is used as the mobile ground platform. It is worth to mention that most of the simulation experiments are performed on the Husarion model due to the higher simulation speed and only for obtaining a policy for the real experiments, the Clearpath Jackal simulation model is used. After training the agent in the simulated platform, the learned policy is transferred directly to the real robot without any extra tuning of the policy parameters or further training using real world samples. 

To evaluate the quality of the transferred policy, different target positions are set for the robot and a comparison is made between the
travelled trajectory on the real and virtual environments. The efficiency of transferring the learned policy has been
validated on different targets where three of them are depicted in Figure \ref{fig:real_robot}. It can be noticed that the trajectories do not perfectly overlap due to the differences that always exist between real world and simulations. Nonetheless, the robot is able to reach the targets without colliding with the obstacles and by following reasonably good trajectories. This proves, once again, the high robustness of learning algorithms against simulation mismatches and noise.

\section{CONCLUSION}

The paper presents a DRL and SLAM map-less path planner for navigation in unknown environments trained by integrating in the reward function the knowledge of map of the training environment, obtained using a grid-based Rao-Blackwellized particle filter (SLAM). To improve learning rate and  accomplish higher success rate in unseen environments, the agent shouldn't only learn the target-reaching tasks, but optimal behaviours in presence of different obstacle configurations, i.e. obstacles awareness. To achieve that, the reward is shaped based on the distance to obstacles in the field of view of the on-board sensor on the robot weighted by the certainty about them given by the map's posterior. The experiments has proven that the proposed approach is not only able to achieve higher convergence speed during training (\textit{Env}-1), 36.9\% improvement, and higher success rate, 25\% higher in \textit{Env}-1, than the standard approach but, the quality of the trajectory is improved thanks to a reduction of the average number of actions, respectively 42.7 less actions  in \textit{Env}-1. Even more importantly, the performance of the agent in unseen environments is drastically enhanced, respectively 23\%, 45\% higher success rate in \textit{Env}-2, \textit{Env}-3 and 64.2, 13.2 less actions on average in  \textit{Env}-2, \textit{Env}-3. Thanks to the incorporation of the obstacle knowledge in the reward function, the  agent  can  better  relate  the  features  extracted from  the  sensor  to  good  or  bad  behaviours,  respectively  getting  closer  to  the  targets  and  getting  closer  to  the obstacles. This proves that the agent has learned how to safely navigate in presence of obstacles. Furthermore, the policy trained in the simulation environment can be successfully transferred to the real robot.




\section*{ACKNOWLEDGMENT}
We thank Klaas Jan Russcher for the precious help and assistance with the Jackal robot. Our experiments benefit from NVIDIA Jetson TX2 which is received as a research grant by Beril Sirmacek.


\bibliographystyle{unsrt}  
\bibliography{main}  

\end{document}